\documentclass[10pt,twocolumn,letterpaper]{article}

\usepackage{wacv} 
\usepackage{graphicx}
\usepackage{amsmath}
\usepackage{amssymb}
\usepackage{multirow}
\usepackage{color}
\usepackage{xcolor}
\usepackage{colortbl}
\usepackage{booktabs}
\usepackage[linesnumbered,lined,vlined,ruled]{algorithm2e}
\usepackage[pagebackref=true, breaklinks=true, colorlinks,
 citecolor=citecolor, linkcolor=linkcolor, bookmarks=false]{hyperref}
\usepackage[capitalize]{cleveref}

\crefname{section}{Sec.}{Secs.}
\Crefname{section}{Section}{Sections}
\Crefname{table}{Table}{Tables}
\crefname{table}{Tab.}{Tabs.}
\definecolor{citecolor}{HTML}{0071BC}
\definecolor{linkcolor}{HTML}{ED1C24}

\def\wacvPaperID{1102} 
\def\confName{WACV}
\def\confYear{2024}
\begin{document}

\title{Wakening Past Concepts without Past Data: \\Class-Incremental Learning from Online Placebos}

\author{Yaoyao Liu$^{1,2}$ 
\quad Yingying Li$^{3}$ 
\quad Bernt Schiele$^{2}$
\quad Qianru Sun$^{4}$\\
\\
\small $^{1}$Johns Hopkins University \quad 
\small $^{2}$Max Planck Institute for Informatics, Saarland Informatics Campus\\
\small $^{3}$University of Illinois Urbana-Champaign\quad  
\small $^{4}$Singapore Management University\\
}
\maketitle

\newcommand{\myparagraph}[1]{\vspace{0.1em}\noindent\textbf{#1}}
\definecolor{mygray}{gray}{0.6}
\definecolor{mygray-bg}{gray}{0.9}
\newcommand{\redt}[1]{\textcolor[rgb]{1,0,0}{#1}}
\newcommand{\whitet}[1]{\textcolor[rgb]{1,1,1}{#1}}
\newcommand{\grayt}[1]{{\textcolor{mygray-bg}{#1}}}

\begin{abstract}

Not forgetting old class knowledge is a key challenge for class-incremental learning (CIL) when the model continuously adapts to new classes. A common technique to address this is knowledge distillation (KD), which penalizes prediction inconsistencies between old and new models. Such prediction is made with almost new class data, as old class data is extremely scarce due to the strict memory limitation in CIL. In this paper, we take a deep dive into KD losses and find that ``using new class data for KD'' not only hinders the model adaption (for learning new classes) but also results in low efficiency for preserving old class knowledge. We address this by ``using the placebos of old classes for KD'', where the placebos are chosen from a free image stream, such as Google Images, in an automatical and economical fashion. To this end, we train an online placebo selection policy to quickly evaluate the quality of streaming images (good or bad placebos) and use only good ones for one-time feed-forward computation of KD. We formulate the policy training process as an online Markov Decision Process (MDP), and introduce an online learning algorithm to solve this MDP problem without causing much computation costs. In experiments, we show that our method 1) is surprisingly effective even when there is no class overlap between placebos and original old class data, 2) does not require any additional supervision or memory budget, and 3) significantly outperforms a number of top-performing CIL methods, in particular when using lower memory budgets for old class exemplars, e.g., five exemplars per class.\footnote{Code: \href{https://github.com/yaoyao-liu/online-placebos}{https://github.com/yaoyao-liu/online-placebos}}
\end{abstract}

\section{Introduction}
\label{sec1_introduction}

AI learning systems are expected to learn new concepts while maintaining the ability to recognize old ones. In many practical scenarios, they cannot access the old data due to the limitations such as storage or data privacy but are expected to be able to recognize all seen classes. A pioneer work~\cite{rebuffi2017icarl} formulated this problem in the class-incremental learning (CIL) pipeline: training samples of different classes are loaded into the memory phase-by-phase, and the model keeps on re-training with new class data (while discarding old class data) and is evaluated on the testing data of both new and old classes. The key challenge is that re-training the model on the new class data tends to override the knowledge acquired from the old classes~\cite{mccloskey1989catastrophic, McRae1993Catastrophic, Ratcliff1990catastrophic,kirkpatrick2017overcoming}, and the problem is called ``catastrophic forgetting''.
To alleviate this problem, most CIL methods~\cite{rebuffi2017icarl,hou2019lucir,douillard2020podnet,liu2020mnemonics,Liu2020AANets,zhu2021class,zhou2022model,wang20233ef,joseph2022energy,zhao2020maintaining,Castro18EndToEnd,kang2022class,mittal2021essentials,Liu2023Continual,Luo2023Class,Liu2023Online,Zhang2023Continual} are equipped with knowledge distillation (KD) losses that penalize any feature and/or prediction inconsistencies between the models in adjacent phases.

\begin{figure*}
\centering
\vspace{-2em}
%\subfigure[Average accuracy]{
\includegraphics[height=2.1in]{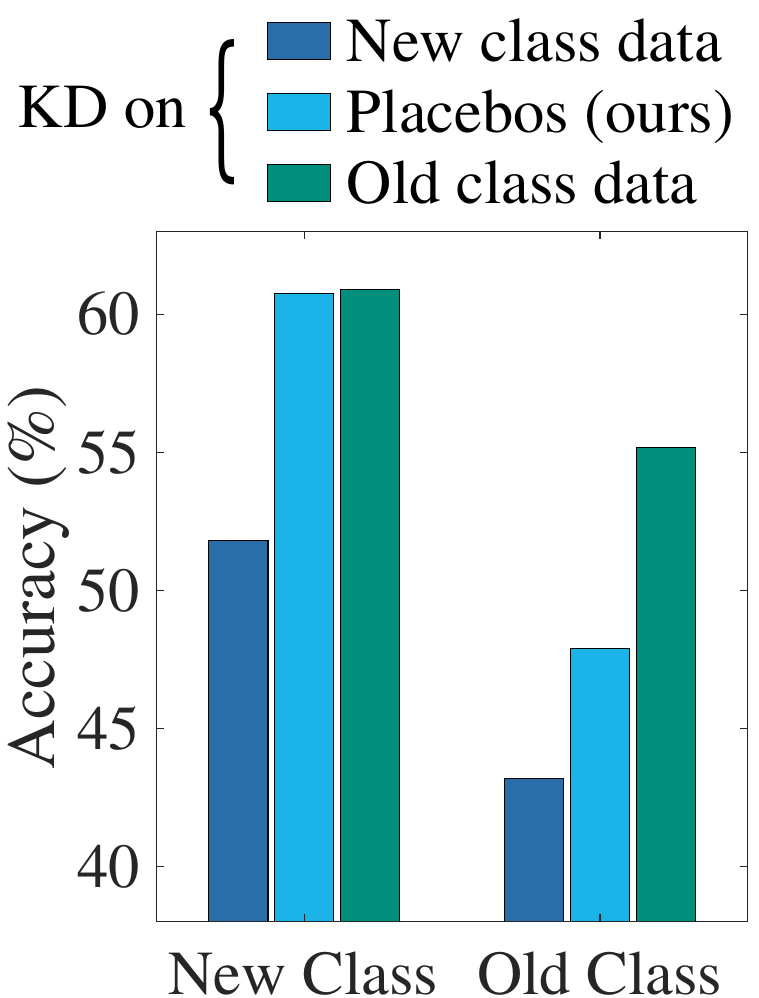}%}
\hspace{0.3cm}
%\subfigure[{Conceptual illustrations of the CIL losses}]{
\includegraphics[height=2.1in]{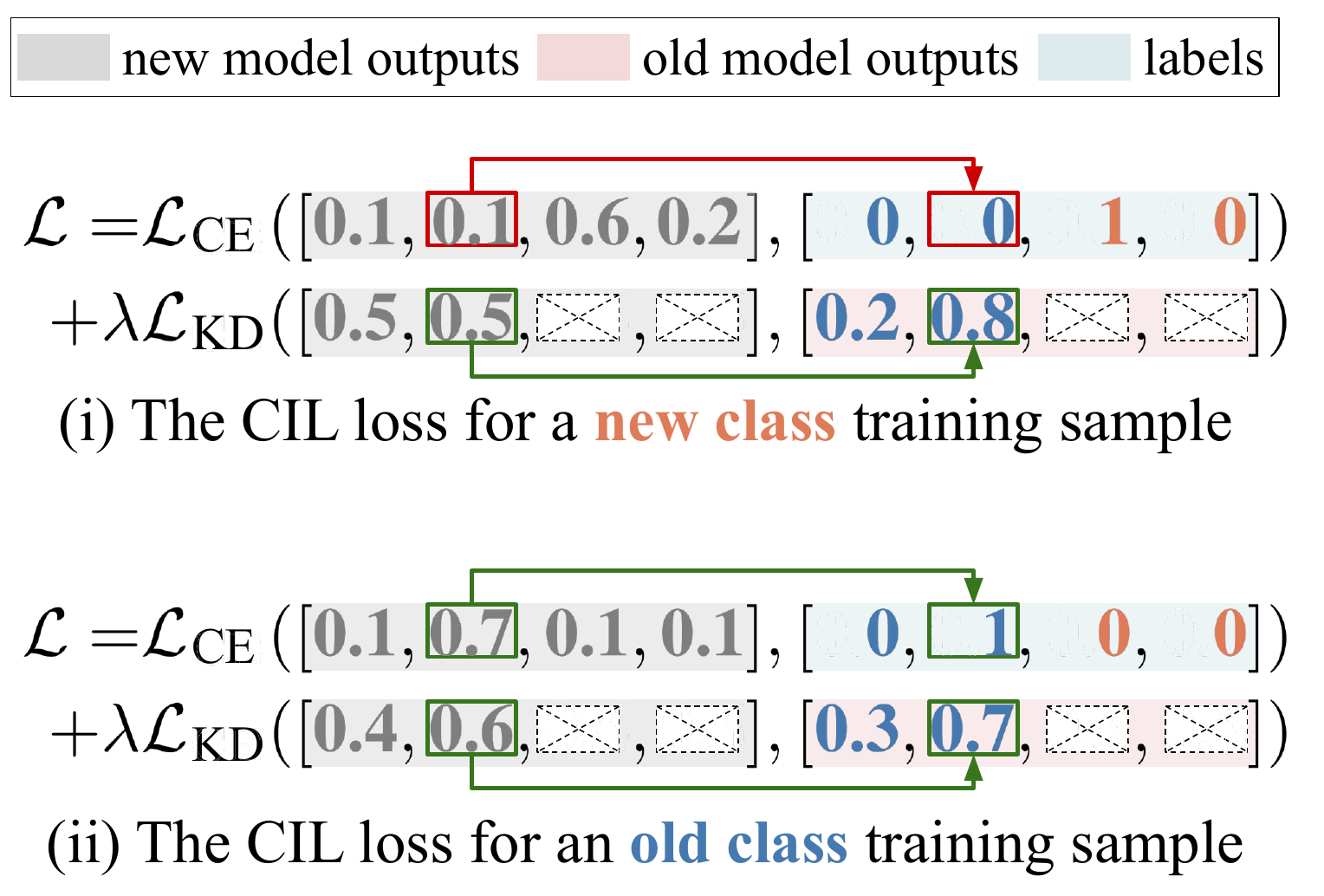}
%}
\hspace{0.3cm}
%\subfigure[Selected placebos]{
\includegraphics[height=2.1in]{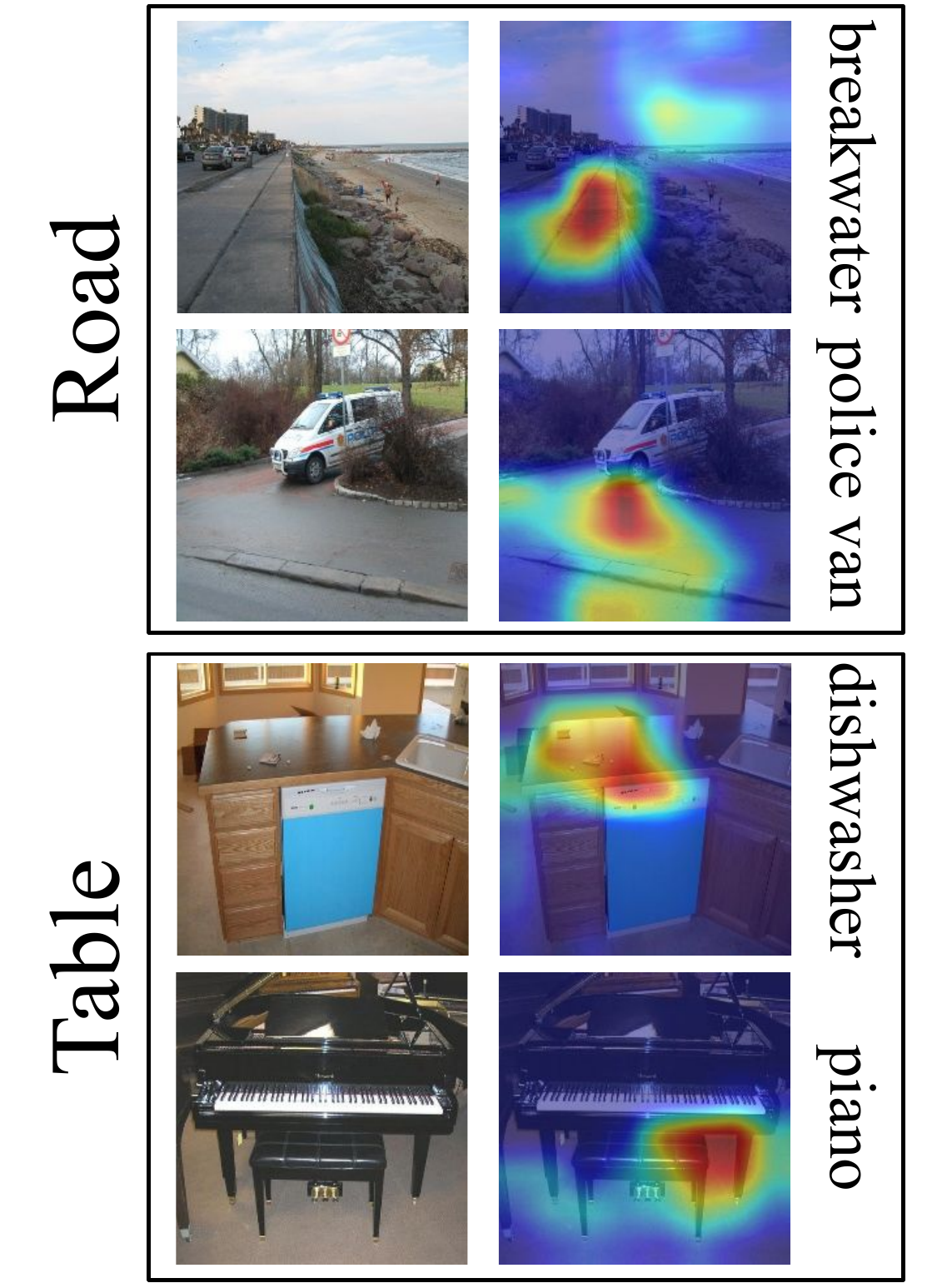}
%}
\vspace{0.5em}
\\
\hspace{0.6cm} (a) Average accuracy \hspace{1.2cm} (b) Conceptual illustrations of the CIL losses \hspace{1.5cm} (c) Selected placebos
\vspace{-0.5em}
\caption{\textbf{(a) 
Average accuracy when computing KD losses on different data} using iCaRL~\cite{rebuffi2017icarl} on CIFAR-100. 
The KD losses (softmax KL divergence loss) are computed on new class data (\textcolor[rgb]{0.276, 0.473, 0.688}{\textbf{dark blue}}), placebos (of old class data) selected by our method (\textcolor[rgb]{0.11,0.71,0.91}{\textbf{light blue}}), and old class data \textcolor[rgb]{0,0.56,0.49}{\textbf{green}}), i.e., the ideal case. 
{\textbf{(b)~Conceptual illustrations of the loss problem if using new class data for KD.} The \textcolor[rgb]{0.276, 0.473, 0.688}{\textbf{dark blue}} and \textcolor[rgb]{0.871, 0.488, 0.351}{\textbf{orange}} numbers denote the predictions of old and new classes, respectively. It is clear in (i) that the objectives are different when using a new class sample for KD (the oracle case is to have both ``\textcolor[rgb]{0.21,0.46,0.11}{\textbf{ascent}}''), e.g., the ground truth label for the second old class is $0$, while the ``KD label'' at this position is $0.8$. This is not an issue when using the old class sample, e.g., in (ii), its ground truth label and ``KD label'' have consistent magnitudes at the same position ($1$ and $0.7$, respectively).}
\textbf{(c)~Our selected placebos for two old classes (``road'' and ``table'') and their activation maps} using GradCAM~\cite{selvaraju2017gradcam} on CIFAR-100. The free image stream is ImageNet-1k which does not have class overlap with CIFAR-100. They are selected because their partial image regions contain similar visual cues to old classes.}
\vspace{-0.5em}
\label{figure_teaser}
\end{figure*}

The ideal KD losses should be computed on old class data since the teacher model (i.e., the model in the last phase) was trained on them. This is, however, impossible in the CIL setting, where almost all old class data are inaccessible in the new phase.
Existing methods have to use new class data as a substitute to compute KD losses. We argue that this 1) hampers the learning of new classes as it distracts the model from fitting the ground truth labels of new classes, and 2) can not achieve the ideal result of KD, as the model can not generate the same soft labels (or features) on new class data as on old class data.
We justify this from an empirical perspective as shown in Figure~\ref{figure_teaser} (a):
the upper bound of KD is achieved when using ``old class data'', and if compared to it, using ``new class data'' sees a clear performance drop for recognizing both old and new classes. In Figure~\ref{figure_teaser} (b), we show the reason by diving into loss computation details: when using new class samples (as substitutes) to compute CE and KD losses simultaneously, these two losses actually weaken each other, which does not happen in the ideal case of using old class samples. 

To solve the above issue, people tried to use unlabeled external data (called \textbf{placebos} in this paper) to compute KD losses (rather than using the new data)~\cite{lee2019overcoming,liu2020more}. 
First, this idea is practical because we don't need to allocate a large memory budget for placebos. We can select a small number of placebos from a free image stream, e.g., Google Image, and delete them immediately after computing KD losses. Second, this idea is effective because computing the KD losses on placebos can help to recall the old class knowledge without weakening the learning of 
new class data. As shown in Figure~\ref{figure_teaser} (a),  compared to the conventional way of using ``new class data'' (for KD), using placebos achieves the same-level new class performance as using ``old class data'', and better old class recognition performance.

However, there are two open questions that need to be addressed when using placebos.
\textbf{Q1}: How to adapt the placebo selection process in the non-stationary CIL pipeline. The ideal selection method needs to handle the dynamics of increasing classes in CIL, e.g., in a later incremental phase, it is expected to handle a more complex evaluation on the placebos of more old classes. \textbf{Q2}: How to control the computational and memory-related costs during the selection and utilization of placebos. It is not intuitive how to process external data without encroaching on the memory allocated for new class data or breaking the strict assumption of memory budget in CIL. Existing works~\cite{lee2019overcoming,zhang2020class} cannot solve the above issues as they use fixed rules for placebo selection and require a large amount of memory to store the placebos.

We solve these questions by proposing a new method called PlaceboCIL that can adjust the policy of selecting placebos for each new incremental phase, in an online and automatic fashion without needing extra memory. Specifically, to tackle \textbf{Q1}, we formulate PlaceboCIL as an online Markov Decision Process (MDP) and introduce a novel online learning algorithm to learn a dynamic policy. In each new phase, this policy produces a phase-specific function to evaluate the quality of incoming placebos. The policy itself gets updated before the next phase.
For \textbf{Q2},
we propose a mini-batch-based memory reusing strategy for PlaceboCIL. Given a free data stream, we sample a batch of unlabeled data, evaluate their quality by using our phase-specific evaluation function (generated by the learned policy), and keep only the high-quality placebos to compute the KD losses.
After this, we remove this batch totally from memory before loading a new batch. In our implementation, this batch can be very small, e.g., $200$ images. \emph{We randomly remove the same size (e.g., $200$) of new class data
to keep the strict assumption of memory budget.}

We evaluate PlaceboCIL by incorporating it into multiple strong baselines such as PODNet~\cite{douillard2020podnet}, LUCIR~\cite{hou2019lucir}, AANets~\cite{Liu2020AANets}, and FOSTER~\cite{wang2022foster}, and conducting a careful ablation study.
Our results on three popular CIL benchmarks show the clear and consistent superiority of PlaceboCIL, especially when using a low memory budget for old class exemplars. For example, our method boosts the last-phase accuracy by $6.9$ percentage points on average when keeping only $5$ exemplars per old class in the memory. 
In addition, it is worth mentioning that \emph{PlaceboCIL is surprisingly efficient even when there is no class overlap between placebos and original old class data.} The reason is that PlaceboCIL can make use of the local visual cues in placebos, e.g., similar visual cues of ``table'' are found on the local regions of an ``piano'' (and ``dishwasher'') image as shown in Figure~\ref{figure_teaser} (c).

\textbf{Our contributions} are three-fold. 1)~A generic PlaceboCIL method that selects placebo images from a free image stream to solve the KD issue in existing methods. 2)~A novel online learning algorithm for training a placebo selection policy and a mini-batch-based memory reusing strategy to avoid extra memory usage. 3)~Extensive comparisons and visualizations on three CIL benchmarks, taking top-performing CIL models as baselines and with the same strict assumption on memory. 

\section{Related Work}
\label{sec2_related_work}

\myparagraph{Class-incremental learning (CIL)} methods can be divided into three categories. \textbf{\emph{{Distillation-based}}} methods introduce different knowledge distillation (KD) losses to consolidate previous knowledge. The key idea is to enforce prediction logits~\cite{Li18LWF,rebuffi2017icarl}, feature maps~\cite{douillard2020podnet}, or other essential information~\cite{Tao2020topology,wang2022foster,simon2021learning,joseph2022energy,yu2020semantic} to be close to those of the pre-phase model. \textbf{\emph{{Memory-based}}} methods use a small number of preserved old class data (called exemplars)~\cite{rebuffi2017icarl,shin2017continual,liu2020mnemonics,prabhu12356gdumb,Luo2023Class,wu2018memory,YanHXHTL022,BangKY0C21,choi2021dual} or augmented data~\cite{zhu2021class} to recall the old class knowledge. \textbf{\emph{{Network-architecture-based}}} methods~\cite{rusu2016progressive,xu2018reinforced,abati2020conditional,yan2021dynamically,Liu2023Online,ZhangSLZPX21}  design incremental network architectures by expanding the network capacity for new class data or freezing partial network parameters to keep the old class knowledge. Our method can be used to improve different \emph{{Distillation-based}} CIL methods.  

Some prior works used unlabeled external data for class-incremental learning. \cite{lee2019overcoming} proposed a confidence-based sampling method to select unlabeled external data to compute a specially designed global distillation loss. \cite{zhang2020class} randomly selected unlabeled samples and used them to compute KD losses for model consolidation. \cite{liu2020more} used unlabeled data to maximize the classifier discrepancy when integrating an ensemble of auxiliary classifiers.  Our method differs from theirs in two aspects. 1) Our method uses the unlabeled data in a more generic way and can be applied to improve different distillation-based methods~\cite{hou2019lucir,rebuffi2017icarl,wang2022foster}, while the existing methods use unlabeled data to assist their specially-designed loss terms or components. 2) We train an online policy to select better-unlabeled data to adapt to the non-stationary CIL pipeline while existing methods select unlabeled data by applying fixed (i.e., non-adaptive) rules in all incremental phases.

\myparagraph{Online learning} observes a stream of samples and makes a prediction for each element in the stream. There are mainly two settings in online learning: {full feedback} and {bandit feedback}. \textbf{\emph{{Full feedback}}} means that the full reward function is given at each stage. It can be solved by Best-Expert algorithms~\cite{even2005experts}. \textbf{\emph{{Bandit feedback}}} means that only the reward of the implemented decision is revealed. If the rewards are independently drawn from a fixed and unknown distribution, we may use, e.g., Thompson sampling~\cite{agrawal2012analysis} and UCB~\cite{auer2010ucb} to solve it. If the rewards are generated in a non-stochastic version, we can solve it by, e.g., Exp3~\cite{auer2002nonstochastic}. \textbf{\emph{{Online MDP}}} is an extension of online learning. Many studies~\cite{even2009online,li2019online2} aim to solve it by converting it to online learning. In our case, we formulate the CIL as an online MDP and convert it into a classic online learning problem. The rewards in our MDP are non-stochastic because the training and validation data change in each phase. Therefore, we design our algorithm based on Exp3~\cite{auer2002nonstochastic}.
\section{Methodology}
\label{sec4_method}
\begin{figure*}
%\vspace{-0.2cm}
\centering
\includegraphics[width=6.8in]{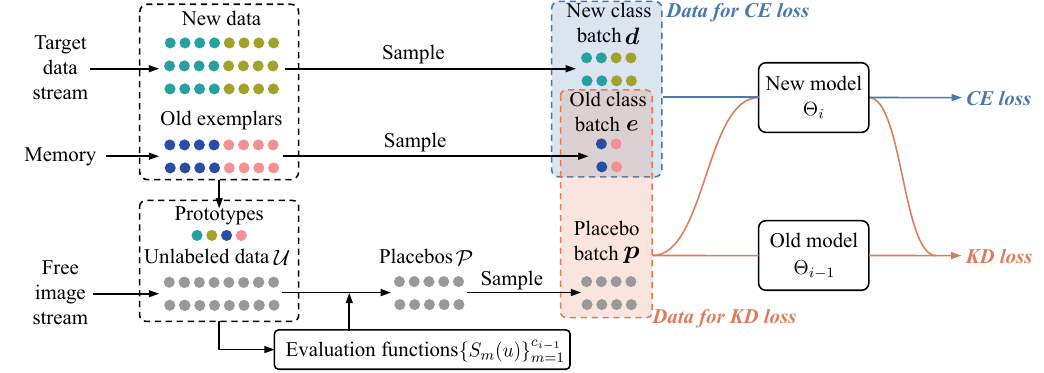}
\vspace{0.0em}
\caption{Our PlaceboCIL in the $i$-th phase. 
At the beginning of this phase, we build phase-specific evaluation functions $\{S_m(u)\}_{m=1}^{c_{i-1}}$. During training, we select placebos as follows. 1)~We load an unlabeled data batch $\mathcal{U}$ from the free image stream. 2)~We compute scores using $\{S_m(u)\}_{m=1}^{c_{i-1}}$ for all samples in $\mathcal{U}$. 3)~For each old class $m$, we select $K$ placebos with the highest scores and add them to $\mathcal{P}$. 4)~We delete used placebos from $\mathcal{P}$ after computing the loss. 5)~When we use up the selected placebos in $\mathcal{P}$, we repeat the selection steps.}
\vspace{-0.3cm}
\label{fig_framework}
\end{figure*}

CIL has multiple ``training-testing'' phases during which the number of classes gradually increases to the maximum. In the $0$-th phase, data $\mathcal{D}_{1:c_0}$=$\{\mathcal{D}_1, ..., \mathcal{D}_{c_0}\}$, including the training samples of $c_0$ classes, are used to learn the model $\Theta_0$. 
After this phase, only a small subset of $\mathcal{D}_{1:c_0}$ (i.e., exemplars denoted as $\mathcal{E}_{1:c_0}$=$\{\mathcal{E}_1, ..., \mathcal{E}_{c_0}\}$) can be stored in the memory and used as replay samples in later phases.
In the $i$-th phase, we use $c_i$ to denote the number of classes we have observed so far. We get new class data $\mathcal{D}_{c_{i-1}+1:c_i}$=$\{\mathcal{D}_{c_{i-1}+1}, ..., \mathcal{D}_{c_i}\}$ of $(c_i-c_{i-1})$ classes and load exemplars $\mathcal E_{1:c_{i-1}}$ from the memory.
Then, we initialize $\Theta_i$ with $\Theta_{i-1}$, and train it using $\mathcal{T}_{1:c_i}$=$\mathcal E_{1:c_{i-1}}${$\cup$}$\mathcal{D}_{c_{i-1}+1:c_i}$. The model $\Theta_{i}$ will be evaluated with a testing set $\mathcal{Q}_{1:c_i}$=$\{\mathcal{Q}_1, ..., \mathcal{Q}_{c_i}\}$ for all classes seen so far. Please note that in any phase of PlaceboCIL, we assume we can access a free image stream, where we can load unlabeled images and select placebos.

PlaceboCIL formulates the CIL task as an online MDP. In each phase, we update a policy, for which we sample a class-balanced subset from training data as the testing set, and use the updated policy to produce a phase-specific evaluation function. During model training, we sample unlabeled images, use the evaluation function to quickly judge the image quality (good or bad placebos), and select the good ones to compute KD losses. In this section, we introduce the formulation of online MDP in Section~\ref{subsec_mdp}, show how to apply the policy to select placebos and compute KD losses in Section~\ref{subsec_online_policy}, and provide an online learning algorithm to update the policy in Section~\ref{subsec_policy_algorithm}. \emph{The pseudocode is given in Algorithms~\ref{algo_Exp3} and \ref{algo_CIL}.}

\subsection{Online MDP Formulation for CIL}
\label{subsec_mdp}

The placebo selection process in CIL should be online inherently: training data (and classes) get updated in each phase, so the placebo selection policy should be updated accordingly. Thus, it is intuitive to formulate the CIL as an online MDP~\cite{Liu2023Online}. In the following, we provide detailed formulations.

\myparagraph{Stages.} Each phase in the CIL task can be viewed as a stage in the online MDP. 

\myparagraph{States.} The state should define the current situation of the agent. In CIL, we use the model $\Theta_i$ as the state of the $i$-th phase (i.e., stage). We use $\mathbb S$ to denote the state space.

\myparagraph{Actions.} We define the action as {$\mathbf{a}_i$}{$=$}$(\beta_i, \gamma_i)$, consisting of the hyperparameters ($\beta_i$ and $\gamma_i$) used to create an evaluation function. As $\beta_i$ and $\gamma_i$ vary in a continuous range, we \emph{discretize} them to define a \emph{finite} action space.\footnote{Though discretization suffers the curse of dimensionality, our experiments show that with a coarse grid, we already have significant improvements over pre-fixed hyperparameters.}
We will elaborate on how to take an action and deploy the hyperparameters in Section~\ref{subsec_online_policy}.

\myparagraph{Policy} {$\pi$}{$=$}$\{p({\mathbf{a}|\Theta_i})\}_{\mathbf{a}\in\mathbb A}$ is a probability distribution over the action space $\mathbb A$, given the current state $\Theta_i$. We will elaborate on how to update the policy using our proposed online learning algorithm in Section~\ref{subsec_policy_algorithm}.

\myparagraph{Environments.} We take the training and testing data in each phase as the environment. In the $i$-th phase, the environment is {$\mathcal{H}_i$}{$=$}$(\mathcal T_{1:c_{i}}, \mathcal Q_{1:c_i})$, where $\mathcal T_{1:c_{i}}$ is the training data and $\mathcal Q_{1:c_i}$ is the testing data. 
The environment is time-varying because we observe different training data (and classes) in each new phase. 

\myparagraph{Rewards.}  CIL aims to train a model that is efficient in recognizing all classes seen so far. Therefore, it is intuitive to use testing accuracy as the reward in each phase. We cannot observe any reward (i.e., testing accuracy) directly because the testing data is not accessible
during training. We solve this by building a local testing set using a subset of training data (see details in Section~\ref{subsec_policy_algorithm}). 
Our objective is to maximize a cumulative reward, i.e., $R=\sum_{i=1}^{N}r_{\mathcal{H}_i}(\Theta_i,\mathbf{a}_i)$, where $r_{\mathcal{H}_i}(\Theta_i,\mathbf{a}_i)$ denotes the $i$-th phase reward.
The reward function $r_{\mathcal{H}_i}$ changes with ${\mathcal{H}_i}$, so it is time-varying.

\subsection{Placebo Selection}
\label{subsec_online_policy}

In the following, we introduce how to build phase-specific evaluation functions using the policy, select high-quality placebos without breaking memory constraints, and compute KD losses with the selected placebos. The computation flow (in each phase) is illustrated in Figure~\ref{fig_framework}.

\myparagraph{Computing prototypes.} Our placebo selection is based on the distance from the placebo to the class prototype, i.e., the mean feature of each class~\cite{snell2017prototypical}. First, we compute the prototypes of all seen classes. We use exemplars to compute the prototypes of old classes, and use new class training data for new class prototypes, as follows,
\begin{equation}
\label{eq_proto}
\begin{aligned}
    \mathrm{Pro}(\mathcal{E}_n) = \frac{1}{|\mathcal{E}_n|}\sum_{z\in\mathcal{E}_n} \mathcal{F}_{\Theta_i}(z), \ \ \mathrm{Pro}(\mathcal{D}_l) = \frac{1}{|\mathcal{D}_l|}\sum_{z\in\mathcal{D}_l} \mathcal{F}_{\Theta_i}(z),
\end{aligned}
\end{equation}
where $\mathcal{F}_{\Theta_i}(\cdot)$ denotes the encoder (i.e., the feature extractor) of $\Theta_i$. $\mathrm{Pro}(\mathcal{E}_n)$ and $\mathrm{Pro}(\mathcal{D}_l)$ denote the prototypes of the $n$-th old class and the $l$-th new class, respectively.

\begin{algorithm}[ht!]%[t]
 % \SetInd{0.000ex}{1.5ex}
 \DontPrintSemicolon
 \SetKwInOut{Input}{Input}
 \SetKwInOut{Output}{Output}
 \vspace{0.05cm}
 \Input{Old model $\Theta_{i-1}$, training data $\mathcal T_{1:c_i}$, testing data $\mathcal Q_{1:c_i}$, learnable parameters $\mathbf w$, numbers of epochs $M_1$ and $M_2$.
 %auxiliary parameters $\{\varrho_{\mathbf{a}}\}_{\mathbf{a}\in\mathbb A}$ in Phase $i$-1 (when $i\geq2$)
 }
 \Output{New model $\Theta_{i}$, new exemplars $\mathcal E_{0:i}$, learnable parameters $\mathbf w$.}
 {\footnotesize \tcp{Policy learning}}
 \If{\rm $i$=$1$}{Initialize $\mathbf w=\{1,\dots,1\}$;}
 %\Else{Initialize $\{\varrho_{\mathbf{a}}\}_{\mathbf{a}\in\mathbb A}$ using the auxiliary parameters learned in Phase $i$-$1$;}
 \For{\rm $t$ in $i, ..., T$}{
 Randomly sample a class-balanced subset $\mathcal B_{1:c_i}$ from $\mathcal T_{1:c_i}$,;\\
 Create the local environment $h_i=((\mathcal T_{1:c_i})\setminus\mathcal B_{1:c_i}, \mathcal B_{1:c_i})$;\\
 Set the policy $\pi={\mathbf w}/{||\mathbf w||}$;\\
 Sample an action $\mathbf{a}_t\sim\pi$;\\
 \For{\rm $j$ in $i, ..., i+n$}{
 Train $\Theta_{j}$ for $M_1$ epochs by \textbf{Algorithm~\ref{algo_CIL}} with inputs $\Theta_{j-1}$, $\mathbf{a}_t$, $h_i$;\\
 Collect the reward $r_{h_i}(\Theta_j, \mathbf{a}_t)$;}
 Compute the cumulative reward ${R}(\mathbf{a}_t, h_i)$ by Eq.~\ref{eq_reward_1};\\
 Update $\mathbf w$ by Eq.~\ref{eq_exp3_update};\\
 }
 {\footnotesize \tcp{CIL training}}
 Sample an action $\mathbf{a}_i\sim\pi$;\\
 Train $\Theta_{i}$ for $M_2$ epochs by \textbf{Algorithm~\ref{algo_CIL}} with inputs $\Theta_{i-1}$, $\mathbf{a}_i$, $\mathcal H_i=(\mathcal T_{1:c_i}, \mathcal Q_{1:c_i})$;\\
 Select new exemplars  $\mathcal E_{1:c_i}$ from $\mathcal T_{1:c_i}$.% using herding~\cite{rebuffi2017icarl}.
 \caption{Our PlaceboCIL in Phase $i$ ({$i$}{$\geq$}{$1$})}
 \label{algo_Exp3}
 \end{algorithm}

 %\vspace{0.3cm}

 \begin{algorithm}[ht!]
 % \SetInd{0.000ex}{1.5ex}
 \DontPrintSemicolon
 \SetKwInOut{Input}{Input}
 \SetKwInOut{Output}{Output}
 \vspace{0.05cm}
 \Input{Old model $\Theta_{\text{old}}$, action $\mathbf{a}=\{\beta, \gamma\}$, environment $h=\{\mathcal T,\mathcal Q\}$.}
 \Output{New model $\Theta$, reward $r_h(\Theta, \mathbf{a})$  (i.e., the testing accuracy).}
 Initialize $\Theta$ with $\Theta_{\text{old}}$;\\
 Create $\{S_m(x)\}_{m=1}^{c_{i-1}}$ based on $\mathbf{a}=\{\beta, \gamma\}$ using Eq.~\ref{eq_evaluation_fucntion};\\
 %{\color[rgb]{0.165, 0.431, 0.686} \footnotesize \tcp{Training}}
 \For{epochs}{
 Set $\mathcal{P}=\varnothing$;\\
 \While{$\mathcal{P}==\varnothing$ }{
Sample $\mathcal{U}$ from the free image stream;\\
Select placebos $\mathcal{P}\subset\mathcal{U}$ using Eq.~\ref{eq_selecting_knockoffs};\\
\For{iterations}{
  Sample mini-batches  $\boldsymbol{p}$, $\boldsymbol{d}$, and $\boldsymbol{e}$;\\
  Compute the loss $\mathcal{L}$ by Eq.~\ref{eq_overall_loss} and update $\Theta$;\\
  Update placbo buffer $\mathcal{P}:=\mathcal{P}\setminus \boldsymbol{p}$;\\
  }
 }
 }
 %{\color[rgb]{0.165, 0.431, 0.686} \footnotesize \tcp{Evaluation}}
 Compute the reward $r_h(\Theta, \mathbf{a})$ on $\mathcal Q$.
 \caption{Training with placebos for action $\mathbf{a}$}
 \label{algo_CIL}
 \end{algorithm}

\myparagraph{Building evaluation functions.} 
We argue that high-quality placebos for the $m$-th old class should meet two requirements: (1) being close to the prototype of the $m$-th class in the feature space because they will be used to activate the related neurons of the $m$-th old class in the model; 
and (2) being far from the prototypes of all the other classes in the feature space so that they will not cause the KD issue (as shown in Figure~\ref{figure_teaser}). 
To achieve these, we design the following evaluation function $\mathcal{S}_m(x)$ for the $m$-th old class in the $i$-th phase:
\begin{equation}
\label{eq_evaluation_fucntion}
    \begin{aligned}
    \mathcal{S}_m(x) 
    =&-{\mathrm{Sim}\left(\mathcal{F}_{\Theta_i}(x),\mathrm{Pro}(\mathcal{E}_m)\right)}\\
    &+\beta_i\sum_{\substack{n=1 \\ n\neq m}}^{c_{i-1}}\frac{\mathrm{Sim}\left(\mathcal{F}_{\Theta_i}(x), \mathrm{Pro}(\mathcal{E}_n)\right)}{c_{i-1}-1}\\
    &+\gamma_i\sum_{l=c_{i-1}+1}^{c_{i}}\frac{\mathrm{Sim}\left(\mathcal{F}_{\Theta_i}(x),\mathrm{Pro}(\mathcal{D}_l)\right)}{c_{i}-c_{i-1}},
    \end{aligned}
\end{equation}
where $x$ denotes an unlabeled input image, 
and $\mathrm{Sim}(\cdot,\cdot)$ denotes cosine similarity. $\beta_i$ and $\gamma_i$ are two hyperparameters from the action {$\mathbf{a}_i$}{$=$}$(\beta_i, \gamma_i)$, sampled by the policy {${\pi}$}.

\myparagraph{Allocating mini-batch-based memory for placebos.} We need to allocate a small amount of memory to store unlabeled images (before evaluating them). At the beginning of the $i$-th phase, we allocate memory buffers $\mathcal{U}$ and $\mathcal{P}$ respectively for the unlabeled image candidates and the selected placebos. \emph{In order to not exceed the memory budget, we randomly remove the same number, i.e.,  $|\mathcal{U}+\mathcal{P}|$, of samples from the training data of new classes.} Our empirical results show this ``remove'' does not degrade the model performance on new classes. 

\myparagraph{Selecting placebos.} Whenever the placebo buffer $\mathcal{P}$ is empty, we load a batch of unlabeled samples $\mathcal{U}$ from the free image stream, and choose $K$ placebos for each old class to add into $\mathcal{P}$, as follows,
\begin{equation}
\label{eq_selecting_knockoffs}
\mathcal{P}: = \{x_k\}_{k=1}^{c_{i\!-\!1}\times K} = {\text{argmax}}_{x_k\in\mathcal{U}}\sum_{m=1}^{c_{i\!-\!1}}\sum_{k=1}^K\mathcal{S}_m(x_k).
\end{equation}
\myparagraph{Calculating loss with placebos.} After selecting placebos, we sample a batch of new class data $\boldsymbol{d}\!\subset\!\mathcal{D}_{c_{i-1}+1:c_i}$, a batch of old class exemplars $\boldsymbol{e}\!\subset\!\mathcal{E}_{1:c_0}$, and a batch of placebos $\boldsymbol{p}\!\subset\!\mathcal{P}$. We calculate the overall loss as follows,
\begin{equation}
\label{eq_overall_loss}
\mathcal{L}=\mathcal{L}_{\mathrm{CE}}(\Theta_i; \boldsymbol{d}\cup \boldsymbol{e}) + \lambda\mathcal{L}_{\mathrm{KD}}(\Theta_{i-1}, \Theta_i; \boldsymbol{p}\cup\boldsymbol{e}),
\end{equation}
where $\mathcal{L}_{\mathrm{CE}}$ and $\mathcal{L}_{\mathrm{KD}}$ denote the CE loss and KD losses, respectively. $\lambda$ is a hyperparameter to balance the two losses~\cite{rebuffi2017icarl}. To control the memory usage, we delete $\boldsymbol{p}$ from $\mathcal{P}$ immediately after calculating the loss. When $\mathcal{P}$ is empty, we repeat the placebo selection operation. 

\setlength{\tabcolsep}{4.1mm}{
\begin{table*}%[ht]
  \small
  \centering
  %\vspace{-0.2cm}
  \begin{tabular}{lcccccccccccc}
  \toprule
  %\\[-11pt]
    \multirow{2.5}{*}{Method} & \multicolumn{2}{c}{$20$ exemplars/class} && \multicolumn{2}{c}{$10$ exemplars/class} && \multicolumn{2}{c}{$5$ exemplars/class}\\
  %\\[-11pt]
  \cmidrule{2-3} \cmidrule{5-6} \cmidrule{8-9}
    %\\[-11pt]
    & Average & Last  && Average & Last && Average & Last \\
      %\\[-11pt]
    \midrule

    LwF~\cite{Li18LWF} & 53.19 \tiny{\whitet{+1.35}} & 43.18 \tiny{\whitet{+3.64}} && 45.96 \tiny{\whitet{+3.64}} & 34.10 \tiny{\whitet{+3.64}} && 35.41  \tiny{\whitet{+13.64}} & 24.91 \tiny{\whitet{+13.64}}\\
        %\\[-11pt]
    \cellcolor{mygray-bg}{\ \ \emph{w/} ours} &  \cellcolor{mygray-bg}{59.08 \tiny{\redt{+5.89}}} & \cellcolor{mygray-bg}{49.15 \tiny{\redt{+5.97}}} & \cellcolor{mygray-bg}{} & \cellcolor{mygray-bg}{53.61 \tiny{\redt{+7.65}}} & \cellcolor{mygray-bg}{38.36 \tiny{\redt{+4.26}}} & \cellcolor{mygray-bg}{} & \cellcolor{mygray-bg}{41.55 \tiny{\redt{+6.14}\grayt{1}}} & \cellcolor{mygray-bg}{28.68 \tiny{\redt{+3.77}\grayt{1}}}\\
        \\[-13pt]
     \cmidrule{1-9}
     
    iCaRL~\cite{rebuffi2017icarl} & 57.12 \tiny{\whitet{+1.35}} & 47.49 \tiny{\whitet{+3.64}} && 53.43  \tiny{\whitet{+3.64}} & 41.49 \tiny{\whitet{+3.64}} && 43.73  \tiny{\whitet{+13.64}} & 34.33 \tiny{\whitet{+13.64}}\\
        %\\[-11pt]
    \cellcolor{mygray-bg}{\ \ \emph{w/} ours} &  \cellcolor{mygray-bg}{61.24 \tiny{\redt{+4.12}}} & \cellcolor{mygray-bg}{51.47 \tiny{\redt{+3.98}}} & \cellcolor{mygray-bg}{} & \cellcolor{mygray-bg}{59.11 \tiny{\redt{+5.68}}} & \cellcolor{mygray-bg}{46.42 \tiny{\redt{+4.93}}} & \cellcolor{mygray-bg}{} & \cellcolor{mygray-bg}{51.55 \tiny{\redt{+7.82}\grayt{1}}} & \cellcolor{mygray-bg}{39.35 \tiny{\redt{+5.02}\grayt{1}}}\\
        \\[-13pt]
    \cmidrule{1-9} 
    
    LUCIR~\cite{hou2019lucir} & 63.17 \tiny{\whitet{+1.35}} & 53.71 \tiny{\whitet{+3.64}} && 60.50  \tiny{\whitet{+3.64}} & 49.08 \tiny{\whitet{+3.64}} && 51.36  \tiny{\whitet{+13.64}} & 39.37 \tiny{\whitet{+13.64}}\\
        %\\[-11pt]
    \cellcolor{mygray-bg}{\ \ \emph{w/} ours} &  \cellcolor{mygray-bg}{65.28 \tiny{\redt{+2.11}}} & \cellcolor{mygray-bg}{56.23 \tiny{\redt{+2.52}}} & \cellcolor{mygray-bg}{} & \cellcolor{mygray-bg}{64.79 \tiny{\redt{+4.29}}} & \cellcolor{mygray-bg}{55.44 \tiny{\redt{+6.36}}} & \cellcolor{mygray-bg}{} & \cellcolor{mygray-bg}{62.74 \tiny{\redt{+11.35}}} & \cellcolor{mygray-bg}{53.25 \tiny{\redt{+13.88}}}\\
        \\[-13pt]
    \cmidrule{1-9}

    LUCIR-AANets~\cite{Liu2020AANets} & 66.72  \tiny{\whitet{+1.35}} & 55.77 \tiny{\whitet{+3.64}} && 61.12  \tiny{\whitet{+3.64}} & 48.83 \tiny{\whitet{+3.64}} && 53.81  \tiny{\whitet{+13.64}} & 42.93  \tiny{\whitet{+13.64}}\\
      %\\[-11pt]
    \cellcolor{mygray-bg}{\ \ \emph{w/} ours} &  \cellcolor{mygray-bg}{67.16 \tiny{\redt{+0.44}}} & \cellcolor{mygray-bg}{59.14 \tiny{\redt{+3.37}}} & \cellcolor{mygray-bg}{} & \cellcolor{mygray-bg}{64.30 \tiny{\redt{+3.18}}} & \cellcolor{mygray-bg}{52.92 \tiny{\redt{+4.09}}} & \cellcolor{mygray-bg}{} & \cellcolor{mygray-bg}{60.27 \tiny{\redt{+6.46}\grayt{1}}} & \cellcolor{mygray-bg}{48.45 \tiny{\redt{+5.52}\grayt{1}}}\\
      \\[-13pt]
    \cmidrule{1-9}
    FOSTER~\cite{wang2022foster} & 70.62 \tiny{\whitet{+1.35}} & 62.97 \tiny{\whitet{+3.64}} && 62.03 \tiny{\whitet{+3.64}} & 52.23 \tiny{\whitet{+3.64}} && 56.80 \tiny{\whitet{+13.64}} & 43.11  \tiny{\whitet{+13.64}}\\
      %\\[-11pt]
    \cellcolor{mygray-bg}{\ \ \emph{w/} ours} &  \cellcolor{mygray-bg}{71.97 \tiny{\redt{+1.35}}} & \cellcolor{mygray-bg}{64.43 \tiny{\redt{+1.46}}} & \cellcolor{mygray-bg}{} & {\cellcolor{mygray-bg}{65.12 \tiny{\redt{+3.09}}}} &\cellcolor{mygray-bg}{{54.81 \tiny{\redt{+2.48}}}} & \cellcolor{mygray-bg}{} & \cellcolor{mygray-bg}{62.78 \tiny{\redt{+5.98}\grayt{1}}} & \cellcolor{mygray-bg}{50.72 \tiny{\redt{+7.61}\grayt{1}}}\\
      \\[-13pt]
      
  \bottomrule

\end{tabular}
\vspace{-0.3em}
  \caption{
  Evaluation results (\%) on CIFAR-100 ($N$=$5$) using different baselines \emph{w/} and \emph{w/o} our PlaceboCIL.
  ``Average'' denotes the average accuracy over all phases. ``Last'' denotes the last phase ($5$-th phase) accuracy.
 }
 \vspace{-0.3cm}
  \label{table_ablation_exemplars}
\end{table*}
}

\subsection{Online Policy Learning Algorithm}
\label{subsec_policy_algorithm}

A common approach to solving an online MDP is to approximate it as an online learning problem and solve it using online learning algorithms~\cite{even2005experts,agrawal2012analysis,auer2002nonstochastic}. We also follow this idea in PlaceboCIL, and our approximation follows~\cite{even2009online}, which is theoretically proved to have the optimal regret. Specifically, Even-Dar et al.~\cite{even2009online} relax the Markovian assumption of the MDP by decoupling the cumulative reward function and letting it be time-dependent so that they can solve online MDP by standard online learning algorithms. 

However, we cannot directly apply the algorithms proposed in \cite{even2009online} to our problem. It is because they assume \textit{full feedback}, i.e., the model can observe the rewards of all actions in every learning phase (which is also why its online learning problem can be solved by Best Expert algorithms~\cite{even2005experts}).
While in CIL, we cannot observe any reward (i.e., the testing accuracies) because the testing data $Q_{1:c_i}$ are not accessible in any phase $i$. 
To address this problem, we split the training data we have in each phase into two subsets: one for training and another for validation. 
Once we have a validation set, we can solve our online learning problem based on Exp3~\cite{auer2002nonstochastic,Liu2023Online}---a simple and effective bandit algorithm.
In the following, we elaborate on how we do this data splitting in each local dataset (i.e., the entire data we have in each training phase of CIL), compute the decoupled cumulative reward, and learn the policy $\pi$ with Exp3.

\myparagraph{Rebuilding local datasets.} 
To compute reward, we sample a class-balanced subset $\mathcal B_{1:c_i}$ from the training data $\mathcal T_{1:c_1}$. 
$\mathcal B_{1:c_i}$ contains the same number of samples for both the old and new classes.
In this way, we rebuild the local training and validate sets, and update the environment from the oracle $\mathcal{H}_i${$=$}{$(\mathcal T_{1:c_{i}}, \mathcal Q_{1:c_i})$} (which is unavailable in CIL) to the local environment $h_i${$=$}{$(\mathcal T_{1:c_{i}}\setminus\mathcal B_{1:c_i}, \mathcal B_{1:c_i})$}.

\myparagraph{Decoupled cumulative reward.} 
We create the decoupled cumulative reward function ${R}$ based on the original cumulative reward function $\sum_{j=1}^{N}r_{{\mathcal{H}}_j}(\Theta_j,\mathbf{a}_j)$. In the $i$-th phase, we compute  ${R}$ as follows,
\begin{equation}
\label{eq_reward_1}  
{R}(\mathbf{a}_i, h_i)={\sum_{j=i}^{i+n}r_{h_i}(\Theta_j,\mathbf{a}_i)}+\text{constant},
\end{equation}
where the ``constant'' denotes the historical rewards from the $1$-st phase to the $(i$-$1)$-th phase. It doesn't influence policy optimization. ${R}(\mathbf{a}_i, h_i)$ is the long-term reward of a time-invariant local MDP based on the local environment $h_i$. We use ${R}(\mathbf{a}_i, h_i)$ as an estimation of the final cumulative reward, following~\cite{even2009online}. Because we don't know the total number of phases $N$ during training, we assume there will be $n$ phases in the future. Furthermore, we fix the action $\mathbf{a}_i$ to simplify the training process. ${R}(\mathbf{a}_i, h_i)$ is a function of $\mathbf{a}_i$ and $h_i$.

\myparagraph{Training policy with Exp3.} 
Exp3~\cite{auer2002nonstochastic} introduces an auxiliary variable $\mathbf w=\{w({\mathbf{a}})\}_{\mathbf{a}\in\mathbb A}$. It is updated as follows.
In the $1$-st phase, we initialize $\mathbf w$ as $\{1,\dots,1\}$. In each phase~$i$ ({$i$}{$\geq$}{$1$}), we update $\mathbf w$ for $T$ iterations. In the $t$-th iteration, we sample an action {$\mathbf{a}_t$}{$\sim$}{$\pi$}, apply the action $\mathbf{a}_t$ to the CIL system,  and compute ${R}(\mathbf{a}_t, h_i)$ using Eq.~\ref{eq_reward_1}. After that, we update $w({\mathbf{a}}_t)$ in $\mathbf w$ as,
\begin{equation}
\label{eq_exp3_update}
w({\mathbf{a}}_t) \gets w({\mathbf{a}_t})\exp(\xi{{R}(\mathbf{a}_t, h_i)}/{p(\mathbf{a}_t|\Theta_i)}),
\end{equation}
where $\xi$ is a constant, which 
can be regarded as the learning rate. 
After updating $\mathbf w$, we get the policy 
$\pi${$=$}${\mathbf w}/{||\mathbf w||}$.
\emph{The pseudocode is available in Algorithms~\ref{algo_Exp3} and \ref{algo_CIL}.}

\section{Experiments}
\label{sec5_experiments}
\newcommand{\highest}[1]{\textbf{#1}}

\begin{table*}%[ht]
  \small
  \centering
  \vspace{-0.3cm}
  \setlength{\tabcolsep}{2.2mm}{
  \begin{tabular}{lccccccccccc}
  \toprule
      %\\[-11pt]
   \multirow{2.5}{*}{Method} & \multicolumn{3}{c}{\emph{CIFAR-100}} && \multicolumn{3}{c}{\emph{ImageNet-100}} && \multicolumn{2}{c}{\emph{ImageNet-1k}}\\
       %\\[-11pt]
  \cmidrule{2-4} \cmidrule{6-8} \cmidrule{10-11}
      %\\[-11pt]
   & $N$=$5$ & $10$  & $25$ && $5$ & $10$ & $25$ && $5$ & $10$ \\
       %\\[-11pt]
    \midrule
    TPCIL~\cite{Tao2020topology} & 65.34 & 63.58 & -- && 76.27 & 74.81 & -- && 64.89 & 62.88  \\
   {GeoDL~\cite{simon2021learning}} &{65.14} &{65.03} & {63.12} &&{76.63} & {75.40} &{71.43} && {65.23} & {64.46} \\
   {DER~\cite{yan2021dynamically}} &{68.65} &{67.48} & {66.18} &&{78.40} & {78.20} &{75.40} && {68.13} & {65.97} \\
     {ELI~\cite{joseph2022energy}} & {68.78} & {66.62} & {64.72} && {73.54} & {71.82} & {70.32} &&  {--} &  {--} \\
    \midrule
        GD+ext~\cite{lee2019overcoming} & 63.17{\tiny{$\pm$0.47}} & 58.71{\tiny{$\pm$0.39}} &  51.79{\tiny{$\pm$0.42}} &&  75.67{\tiny{$\pm$0.51}} &  72.08{\tiny{$\pm$0.61}} &  65.13{\tiny{$\pm$0.56}} &&  -- &  --  \\
    MUC-LwF~\cite{liu2020more} & 59.03{\tiny{$\pm$0.35}} & 53.27{\tiny{$\pm$0.47}} &  49.06{\tiny{$\pm$0.49}} &&  72.31{\tiny{$\pm$0.53}} &  68.92{\tiny{$\pm$0.60}} &  62.93{\tiny{$\pm$0.62}} &&  -- &  -- \\
        %\\[-11pt]
    \midrule
    %\\[-11pt]
   {POD-AANets~\cite{Liu2020AANets}}  & {66.12}\tiny{$\pm$0.41} & {64.11}\tiny{$\pm$0.32} &  {62.12}\tiny{$\pm$0.51} &{}& {76.63}\tiny{$\pm$0.47} &  {{75.40}}\tiny{$\pm$0.36} &  {71.43}\tiny{$\pm$0.32} &{}& 67.60\tiny{$\pm$0.39} &  64.79\tiny{$\pm$0.42} \\
   \cellcolor{mygray-bg}{\ \ \emph{w/} PlaceboCIL (ours)}  & \cellcolor{mygray-bg}{{67.65}}\tiny{$\pm$0.45} & \cellcolor{mygray-bg}{{65.78}}\tiny{$\pm$0.40} &  \cellcolor{mygray-bg}{{64.95}}\tiny{$\pm$0.46}&\cellcolor{mygray-bg}{}& \cellcolor{mygray-bg}{{{78.24}}}\tiny{$\pm$0.52} &  \cellcolor{mygray-bg}{{{77.14}}}\tiny{$\pm$0.47} &  \cellcolor{mygray-bg}{{\highest{75.85}}}\tiny{$\pm$0.42} &\cellcolor{mygray-bg}{}& \cellcolor{mygray-bg}{{68.55}}\tiny{$\pm$0.34} & \cellcolor{mygray-bg}{{65.49}}\tiny{$\pm$0.38}  \\
   {FOSTER~\cite{wang2022foster}} &{70.62}\tiny{$\pm$0.58} &{68.43}\tiny{$\pm$0.45} & {63.83}\tiny{$\pm$0.62} &&{80.21}\tiny{$\pm$0.67} & {77.63}\tiny{$\pm$0.73} &{69.27}\tiny{$\pm$0.50} && {69.32}\tiny{$\pm$0.47} & {66.07}\tiny{$\pm$0.61}  \\
   \cellcolor{mygray-bg}{\ \ \emph{w/} PlaceboCIL (ours)}  & \cellcolor{mygray-bg}{\highest{71.97}}\tiny{$\pm$0.49} & \cellcolor{mygray-bg}{\highest{70.31}}\tiny{$\pm$0.59} &  \cellcolor{mygray-bg}{\highest{67.02}}\tiny{$\pm$0.65} &\cellcolor{mygray-bg}{}& \cellcolor{mygray-bg}{{\highest{82.03}}}\tiny{$\pm$0.49} &  \cellcolor{mygray-bg}{{\highest{79.52}}}\tiny{$\pm$0.60} &  \cellcolor{mygray-bg}{{{72.79}}}\tiny{$\pm$0.45} &\cellcolor{mygray-bg}{}& \cellcolor{mygray-bg}{\highest{71.02}}\tiny{$\pm$0.39}& \cellcolor{mygray-bg}{\highest{68.82}}\tiny{$\pm$0.54}  \\
    \\[-13pt]
  \bottomrule

\end{tabular}
}
 \vspace{-0.5em}
  \caption{
  Average accuracy (\%) across all phases. The first block shows top-performing CIL methods. The second block shows  CIL methods that use unlabeled data. The third block shows our method. 
}
%\vspace{-0.5cm}
  \label{table_sota}
\end{table*}

\begin{table}%[ht]
  \small
  \centering
  \setlength{\tabcolsep}{1.20mm}{
  %\vspace{-0.2cm}
 \begin{tabular}{llccccccccccc}
  \toprule
      %\\[-11pt]
    \multirow{2.5}{*}{No.}&\multirow{2.5}{*}{Setting} & \multicolumn{2}{c}{iCaRL} && \multicolumn{2}{c}{LUCIR-AANets} \\
        %\\[-11pt]
  \cmidrule{3-4} \cmidrule{6-7} 
      %\\[-11pt]
   && Average & Last  && Average & Last \\
       %\\[-11pt]
    \midrule
        %\\[-11pt]
    1&Baseline & 57.12   & 47.49  && 66.72  & 57.77  \\
          %\\[-11pt]
    2&PlaceboCIL  & 61.01 & 51.45 && 67.16 & 59.14 \\
        %\\[-11pt]
    \midrule
        %\\[-11pt]
    3&Overlapping & 62.15   & 52.62 && 67.48   & 59.06  \\
          %\\[-11pt]
    4&Non-overlapping & 61.52   & 51.70  && 67.01 & 58.53  \\
          %\\[-11pt]
    5&New data & 57.70  & 47.51  && 66.69 & 57.33  \\
          %\\[-11pt]
    6&Old data (oracle)& 66.64 & 58.03 && 68.82 & 61.52 \\
        %\\[-11pt]
    \midrule
        %\\[-11pt]
    7&\emph{w/o} Online learning  & 60.27 & 50.57  && 66.91 & 58.88  \\
          %\\[-11pt]
    8&Offline RL& 61.09   & 50.81 && 67.31 &59.26  \\

    \midrule
        %\\[-11pt]
    {9}&{Higher confidence} &{60.43}&{49.36}&&{66.97}&{58.12}  \\
    {10}&{Random placebos} &{56.27}&{46.64}&&{66.23}&{57.22}  \\
    %\\[-11pt]

  \bottomrule

\end{tabular}
}
	%}
		%\end{minipage}
		\centering
  \vspace{-0.5em}
	\caption{Ablation results (\%) on CIFAR-100, $N$=$5$. {(1)~\emph{\textbf{First block: baselines.}} Row 1 shows the baselines. Row 2 shows our method. All other settings (Rows 3-10) are based on Row~2. (2)~\emph{\textbf{Second block: different free data streams.}} Rows 3-6 show the ablation results for the following free data streams. 
 (3)~\emph{\textbf{Third block: different policy learning methods.}} Row 7 is for using fixed evaluation functions ($\beta_i$=$\gamma_i$=$1$). Row 8 uses the offline RL (the REINFORCE algorithm) to train the selection policy. (4)~\emph{\textbf{Fourth block: different placebo selection strategies.}} Row 9 uses unlabeled data with higher confidence. Row 10 uses them randomly.
 }
 }
	\label{table_ablation_2}
   \vspace{-0.3cm}

\end{table}

We evaluate our method on three CIL benchmarks and achieve consistent improvements over multiple baseline methods. Below we introduce datasets and implementation details, followed by results and analyses, including the comparison to the state-of-the-art, an ablation study, and the visualization of our placebos.

\myparagraph{Datasets and free image streams.}
We use three datasets: CIFAR-100~\cite{krizhevsky2009learning}, ImageNet-100~\cite{rebuffi2017icarl}, and ImageNet-1k~\cite{russakovsky2015imagenet}. ImageNet-100, which contains $100$ classes, is sampled from ImageNet-1k. We use exactly the same classes or orders as the related works~\cite{rebuffi2017icarl,hou2019lucir}.
For CIFAR-100, we use ImageNet-1k as the free image stream.
For ImageNet-100, we use a $900$-class subset of ImageNet-1k, 
which is the complement of ImageNet-100 in ImageNet-1k. For ImageNet-1k, we use a $1,000$-class subset of ImageNet-21k~\cite{deng2009imagenet} without any overlapping class (different super-classes from those in ImageNet-1k).

\myparagraph{Implementation details.} Following \cite{hou2019lucir,douillard2020podnet,Liu2020AANets,Liu2021RMM}, we use a modified $32$-layer ResNet for CIFAR-100 and an $18$-layer ResNet for ImageNet datasets. 
The number of exemplars for each class is $20$ in the default setting. The training batch size is $128$. 
{On CIFAR-100 (ImageNet-Subset/1k), we train it for $160$ ($90$) epochs in each phase, and divide the learning rate by $10$ after $80$ ($30$) and then after $120$ ($60$) epochs. 
If the baseline is POD-AANets~\cite{Liu2020AANets}, we fine-tune the model for $20$ epochs using only exemplars.}
{We apply different forms of distillation losses on different baselines: (1) if the baselines are LwF and iCaRL, we use the softmax KL divergence loss; (2) if the baselines are LUCIR and AANets, we use the cosine embedding loss~\cite{hou2019lucir}; and (3) if the baseline is POD-AANets, we use pooled outputs distillation loss~\cite{douillard2020podnet}.}
For our PlaceboCIL, $|\mathcal{U}|$ and $|\mathcal{P}|$ are set as $1,000$ and $200$, respectively. All experiments of our PlaceboCIL use the ``strict budget'' setting, i.e., deleting  $|\mathcal{U}+\mathcal{P}|$ samples from training data to avoid exceeding the memory budget. 

\myparagraph{Results on five baselines.}
Table~\ref{table_ablation_exemplars} shows the average and last-phase accuracy for five baselines (i.e., LwF~\cite{Li18LWF}, iCaRL~\cite{rebuffi2017icarl}, LUCIR~\cite{hou2019lucir}, AANets~\cite{Liu2020AANets}, and FOSTER~\cite{wang2022foster}). From the table, we make the following observations. 
1)~Using our PlaceboCIL boosts the performance of the baselines clearly and consistently in all settings, indicating that our method is generic and efficient.
2)~When the number of exemplars decreases, the improvement brought by our method becomes more significant. For example, the last-phase accuracy improvement of LUCIR increases from $2.52$ to $13.88$ percentage points when the number of exemplars per class decreases from $20$ to $5$. This reveals that the superiority of our method is more obvious when the forgetting problem is more serious (with fewer exemplars) due to a tighter memory budget in CIL.
3) Our PlaceboCIL can boost the performance of all KD terms, i.e., not only for logits-based KD~\cite{rebuffi2017icarl} but also for feature-based KD~\cite{hou2019lucir,douillard2020podnet}.

%\begin{wrapfigure}{l}{0.45\textwidth}
\begin{figure*}
\vspace{-0.1cm}
  \begin{center}
  \centering
\includegraphics[width=6.8in]{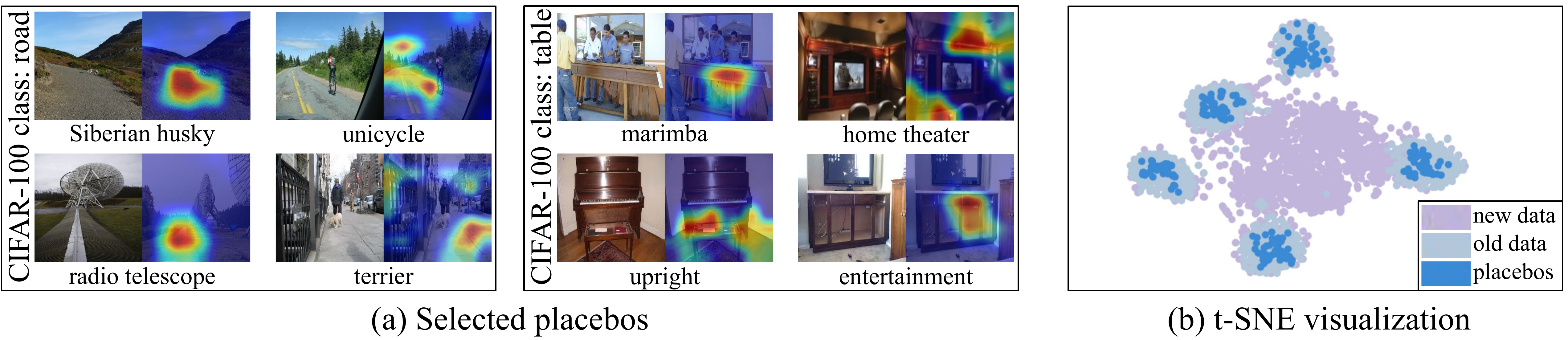}
  \end{center}
  \vspace{-0.6cm}
\caption{(a) Selected placebos for two CIFAR-100 classes and their GradCAM activation maps. The free image stream is non-matching ImageNet-1k. (b) The t-SNE results on CIFAR-100 ($N$=$5$). For clear visualization, we randomly pick five new classes and five old classes. The \textcolor[rgb]{0.75, 0.70, 0.85}{\textbf{purple}}, \textcolor[rgb]{0.66, 0.79, 0.90}{\textbf{light blue}}, and \textcolor[rgb]{0.276, 0.473, 0.688}{\textbf{dark blue}} points denote the new data, old data, and selected placebos, respectively.
}
\label{figure_visualization}
\vspace{-0.2cm}
\end{figure*}
%\end{wrapfigure}

\begin{table}%[ht]
  \small
  \centering
  \setlength{\tabcolsep}{4.30mm}{
  %\vspace{-0.2cm}
 \begin{tabular}{llccccccccccc}
  \toprule
      %\\[-11pt]
    $|\mathcal{U}|$ & $2000$ & $1000$ & $500$ &
    $0$\\
        %\\[-11pt]
       %\\[-11pt]
    \midrule
        %\\[-11pt]
    Acc. (\%) & 61.13 & 61.01   & 58.23 &  
    57.12  \\
    %\\[-11pt]

  \bottomrule

\end{tabular}
}
	%}
		%\end{minipage}
		\centering
  \vspace{-0.5em}
	\caption{Ablation results (\%) for different memory buffer sizes $|\mathcal{U}|$ on CIFAR-100, $N$=$5$. The baseline is iCaRL~\cite{rebuffi2017icarl}.
 }
	\label{table_ablation_u}
   \vspace{-0.3cm}

\end{table}

\myparagraph{Comparisons to the state-of-the-art.} 
Table~\ref{table_sota} (Blocks 1\&3) shows the results of our best model (taking PlaceboCIL as a plug-in module in the top method~\cite{wang2022foster}) and some recent top-performing methods. We can see that using our PlaceboCIL outperforms all previous methods. Intriguingly, we find that we can surpass others more when the number of phases is larger---where there are more serious forgetting problems. For example, when $N$=$25$, we improve POD-AANets by $4.4\%$ on the ImageNet-100, while this number is only $1.6\%$ when $N$=$5$ (which is an easier setting with more saturated results). This reflects the encouraging efficiency of our method for reducing the forgetting of old class knowledge in CIL models.

\myparagraph{Comparisons to the CIL methods using unlabeled data.} Table~\ref{table_sota} (Blocks 2\&3) shows the results of our best model and  CIL methods using unlabeled data (GD+ext~\cite{lee2019overcoming} and MUC-LwF~\cite{liu2020more}). We can see that our method consistently performs better than others. For another related work, DMC~\cite{zhang2020class}, we didn't find the public code. So, we compare ours with DMC using their paper's setting: iCaRL w/ ours achieves $62.3\%$, while the result of DMC is $59.1\%$ (CIFAR-100, 10 phases, 10 classes/phase). 

\myparagraph{Ablation study.} Table~\ref{table_ablation_2} shows the ablation results. 

\noindent\emph{\textbf{1) First block.}} 
Rows 1 and 2 show the baseline and our method, respectively.

\noindent\emph{\textbf{2) Second block: different free data streams.}} Rows 3-6 show the ablation results for the following free data streams. (1) ``Overlapping'' means including samples from the overlapping classes between CIFAR-100 and ImageNet. (2) ``Non-overlapping'' means using only the samples of non-overlapping classes between CIFAR-100 and ImageNet (more realistic than ``Overlapping''). (3) ``New data'' means using only the current-phase new class data (i.e., without using any free data stream) as candidates to select placebos. (4) ``Old data'' means the original old class data are all accessible when computing KD losses (i.e., the upper bound of KD effect). Please note that in (1) and (2), two classes are considered ``overlapping'' if their classes or super-classes overlap. For example, ``n02640242 - sturgeon'' in ImageNet-1k is regarded as an overlapping class of the ``fish'' in CIFAR-100, because they overlap at the level of super-class (i.e., ``fish''). 
When comparing Row 4 with Row 2, we can find that our method is robust to the change of data streams: even if all overlapping classes are removed, our method can still achieve the same-level performance. Comparing Row 5 with Row 2, we can get a clear sense that using additional unlabeled data is definitely helpful.
Comparing Row 6 with Row 2, we see that our method achieves comparable results to the upper bound.

\noindent\emph{\textbf{3) Third block: different policy learning methods.}} {Row 7 is for using fixed evaluation functions ($\beta_i$=$\gamma_i$=$1$).} Row 8 uses the offline RL (the REINFORCE algorithm~\cite{Liu2021RMM}) to train the selection policy. 
Comparing Row 7 with Row 2 shows that using online learning successfully boosts the model performance. Comparing Row 8 with Row 2, we are happy to see that our online learning method achieves the same-level performance as the offline RL while the training time is much less. The training time of the baseline (without learning a policy) is $2.7$ hours.
It becomes around $650$ hours if we solve the MDP by offline RL. In contrast, using our online method takes only $4.5$ hours. 

\noindent{\emph{\textbf{4) Fourth block: different placebo selection strategies.}}} Row 9 uses unlabeled data with higher confidence following~\cite{lee2019overcoming}. 
Row 10 uses them randomly following~\cite{zhang2020class}.
Comparing these results with Row 2 shows our superiority. {The ``mini-batch-based memory reusing strategy'' is applied in Rows 9 and 10.} 

\noindent{\emph{\textbf{5) Different memory buffer sizes.}}} Table~\ref{table_ablation_u} shows the ablation results when using different buffer sizes for $\mathcal{U}$. We can observe that larger buffer sizes achieve better results. Interestingly, we can also see that using a relatively small buffer size (e.g., $500$) can still improve the baseline. 

\myparagraph{Visualization results.} Figure~\ref{figure_visualization} (a) demonstrates
the activation maps visualized by Grad-CAM for the placebos of two old classes on CIFAR-100 (``road'' and ``table''). ImageNet-1k is the free data stream. We can see that the selected placebos contain the parts of ``road'' and ``table'' even though their original labels (on ImageNet-1k) are totally different classes. While this is not always the case, our method seems to find sufficiently related images to old classes that activate the related neurons for old classes (``road'' and ``table''). To illustrate that, Figure~\ref{figure_visualization}~(b) shows t-SNE results for placebos, old class data (not visible during training), and new class data. We can see that the placebos are located near the old class data and far away from the new class data. This is why placebos can recall the old knowledge without harming the new class learning.
\section{Conclusions}
\label{sec_conclusions}
We proposed a novel method, PlaceboCIL, which selects high-quality placebo data from free data streams and uses them to improve the effect of KD in CIL. We designed an online learning method to make the selection of placebos more adaptive in different phases and a mini-batch-based memory-reusing strategy to control memory usage. Extensive experimental results show that our method is general and efficient.

{\small
\bibliographystyle{ieee_fullname}
\bibliography{egbib}
}

\end{document}